\documentclass[
]{ceurart}

\usepackage[utf8]{inputenc}

\usepackage{listings}
\usepackage{comment}
\usepackage{longtable}
\usepackage{appendix}
\usepackage{amsmath}
\usepackage{booktabs}
\usepackage{enumitem}
\usepackage{amssymb}
\usepackage{placeins}
\usepackage{needspace}
\usepackage{hyperref}
\usepackage{tabularx}

\newcommand{\ld}[1]{\textcolor{purple}{Comment by Laura: #1}}
\renewcommand{\ld}[1]{ }

\usepackage[most]{tcolorbox}
\usepackage{xcolor}

\tcbset{
  mypromptstyle/.style={
    enhanced,
    breakable, 
    colback=blue!5!white,
    colframe=blue!50!black,
    boxrule=0.5pt,
    arc=2pt,
    left=4pt,
    right=4pt,
    top=2pt,
    bottom=2pt,
    fontupper=\small\ttfamily
  }
}

\newcommand{\textprompt}[1]{%
  \tcbox[mypromptstyle, on line, box align=base]{#1}%
}

\newtcolorbox{blockprompt}[1][]{mypromptstyle, title=#1}

\tcbset{
  myexamplestyle/.style={
    enhanced,
    breakable, 
    colback=green!5!white,
    colframe=green!50!black,
    boxrule=0.5pt,
    arc=2pt,
    left=4pt,
    right=4pt,
    top=2pt,
    bottom=2pt,
    fontupper=\small\ttfamily
  }
}

\newcommand{\textexample}[1]{%
  \tcbox[myexamplestyle, on line, box align=base]{#1}%
}

\newtcolorbox{blockexample}[1][]{myexamplestyle, title=#1}

\begin{document}

\copyrightyear{2025}
\copyrightclause{Copyright for this paper by its authors.
  Use permitted under Creative Commons License Attribution 4.0
  International (CC BY 4.0).}

\conference{CLEF 2025 Working Notes, 9 -- 12 September 2025, Madrid, Spain}



\title{UNH at CheckThat! 2025: Fine-tuning Vs Prompting in Claim Extraction}
\date{Computer Science Department, University of New Hampshire, USA.}

\author[1]{Joe Wilder}[%
email=Joe.Wilder@unh.edu,%
]\author[1]{Nikhil Kadapala}[%
email=Nikhil.Kadapala@unh.edu,%
]\author[1]{Benji Xu}[%
email=Yanjie.Xu@unh.edu,%
]\author[1]{Mohammed Alsaadi}[%
email=Mohammed.Alsaadi@unh.edu,%
]\author[1]{Aiden Parsons}[%
email=Aiden.Parsons@unh.edu,%
]\author[1]{Mitchell Rogers}[%
email=Mitchell.Rogers@unh.edu,
]\author[1]{Palash Agrawal}[%
email=Palash.Agrawal@unh.edu,%
]\author[1]{Adam Hassick}[%
email=Adam.Hassick@unh.edu,%
]\author[1]{Laura Dietz}[%
email=dietz@cs.unh.edu,%
]

\address[1]{University of New Hampshire,
  Durham, NH, 03824, USA}

\begin{abstract}
We participate in CheckThat! Task 2 English and explore various methods of prompting and in-context learning, including few-shot prompting and fine-tuning with different LLM families, with the goal of extracting check-worthy claims from social media passages. Our best METEOR score is achieved by fine-tuning a FLAN-T5 model. However, we observe that higher-quality claims can sometimes be extracted using other methods, even when their METEOR scores are lower.
\end{abstract}

\maketitle

\section{\texorpdfstring{Introduction
}{Introduction }}\label{introduction}
The CheckThat!~Lab aims to build computational infrastructure to support human fact-checkers, consisting of a pipeline for verifying the truthfulness of social media text.

We participate in Task 2 (English) \cite{clef-checkthat:2025:task2}, which addresses a specific challenge in this pipeline:

\paragraph{Task Statement} Given the text of a social media post, extract the main claim in succinct and concise language suitable for a human fact checker to verify.

\bigskip

We anticipated a number of unique challenges in this task and designed our methods to study them empirically. Many of our concerns align with \citet{thorne2018fever}.

\paragraph{Multiple equally relevant claims.} Often, multiple extracted claims are topically relevant. However, the task requires focusing on a single fact for manual verification. Analyzing the training and validation data, we found that many gold claims include multiple facts and could be rewritten more succinctly. For example: "Joe Biden lives in a large estate bought on a senator's salary."

\paragraph{Missing multi-modal content.} We found that many gold claims refer to information not available to participants, such as photos or videos included with the original post but removed during pre-processing. While we accept that claims cannot be based on unseen data, we instruct our models to make educated guesses about the topic.

\begin{blockexample}
Russia vs Ukraine war 

https://www.huobi.com/en-us/topic/double-invite/register/?
invite\_code=ije73223\&name=BlackWidow\&avatar=6\&inviter\_id=11343840

BREAKING NEWS LIVE 

Latest stars: Martin Johnson -200 

Latest Supporter: Emeka Efobi
\end{blockexample}

\paragraph{Faithful extractions.} We are especially concerned about extracted claims where LLM hallucinations introduce content not present in the original post. An example of a hallucinated and overly verbose claim follows:

\Needspace{2\baselineskip}

\paragraph{Social Media Post:}
\begin{blockexample}
    The salary of a U.S. Senator is \$174,000 per year. This is Joe Biden's
house\ldots{} seems legit
:) The
salary of a U.S. Senator is \$174,000 per year. This is Joe Biden's
house\ldots{} seems legit
:) The
salary of a U.S. Senator is \$174,000 per year. This is Joe Biden's
house\ldots{} seems legit
 :) 
\end{blockexample}

\Needspace{2\baselineskip}
\paragraph{Hallucinated Extracted Claim} 

\bigskip

\begin{blockexample}
Joe Biden's house, purchased for an amount significantly exceeding the cumulative value of his annual U.S. Senator's salary of \$174,000, raises questions about potential additional, undisclosed sources of income that may have contributed to the down payment, mortgage payments, property taxes, insurance premiums, and ongoing maintenance costs associated with the property.
\end{blockexample}

\bigskip

Our emphasis is on exploring the design space across different LLMs and methods---fine-tuning and few-shot prompting---in search of the best trade-off between optimizing the METEOR score and producing claims that are genuinely useful for human fact-checkers.

We conduct a broad exploration of methods, prompts, and LLMs, casting a wide net. Our approaches fall into three overarching categories:

\begin{enumerate}
\item Fine-tuning approaches,
\item Prompting approaches, and
\item "Frustratingly easy" baselines.\footnote{We did not use baselines provided by the organizer.}
\end{enumerate}

We describe all explored approaches and submit those performing best on the validation set in terms of METEOR.

We only use resources provided by the Task 2 organizers and publicly available large language models from Hugging Face and the Together.AI API service.

\section{Approaches: Fine-Tuning and Prompting}
\label{approaches}
In this section, we describe methods relying on fine-tuning across LLMs of different parameter scales.

Our key takeaway: Flan-T5 Large \cite{google-flan-t5} offered the best compromise between raw capability and practical fine-tuning feasibility under our hardware and time constraints.

\subsection{Finetuned Flan-T5 Large\ld{(Joe) - A1}}
\label{finetuned-flan-t5-large-joe}

This approach fine-tuned the Flan-T5 Large \cite{google-flan-t5} model on the CLEF 2025 Task 2 training dataset to align its outputs more closely with the gold-standard claims. Fine-tuning was performed using the \texttt{huggingface} transformers library without advanced techniques such as LoRA or PEFT.

Due to resource limitations, billion-parameter models were out of scope. We opted for Flan-T5 Large (783M parameters), which can run locally and is more manageable to train due to its smaller size. A straightforward task-specific prompt was prepended to training examples:

\begin{blockprompt}
Please read the following social media post and extract the claim made within it.
Normalize the claim by rephrasing it in a clear and concise manner.

Post: \$text

Extracted Claim:
\end{blockprompt}

The training ran for 10 epochs on an NVIDIA 4060 GPU, taking nearly four days to complete. This approach's strength lies in its ability to internalize extraction patterns not easily expressible via prompting alone. It achieved an average validation-set METEOR score of 0.5569.

\subsection{2.1 LoRA fine-tuning of Flan-T5 Base\ld{(Mohammed - A2)}}
\label{t5--base-fine-tuned-using-lora-mohammed}
\label{mohammed}

The motivation behind this submission was to balance performance with efficiency. Full fine-tuning of T5-Base \cite{google-flan-t5} gives strong results but incurres high computational costs. Prompt tuning, while efficient, yields limited gains---especially on larger models. LoRA \cite{hu2021lora} provides a middle ground by updating only ~0.4\% of parameters, keeping the rest of the model frozen.

LoRA allowed us to adapt the model effectively with minimal overhead. We chose Flan-T5 Base due to its strong baseline performance, aiming to retain quality while reducing resource demands.

This model achieved a validation METEOR score of 0.3054 and was our third-best test-set run, with a test METEOR of 0.28.

\subsection{\ld{Adam - A3} Fine-tuned DeepSeek-R1-Distill-Llama-8b Approach}
\label{fine-tuned-llama-3.1-8b-approach} 
\label{adam}

Here, we fine-tuned the \texttt{DeepSeek-R1-Distill-Llama-8b} \cite{deepseek-r1-distill-llama-8b} causal language model on the training dataset to assess its ability to produce reference-style summaries. The training used the following system prompt:

\begin{blockprompt}
Extract the verifiable claim as one sentence from the user input.
\end{blockprompt}

This approach achieved a validation METEOR score of 0.2541.

Fine-tuning yields marginal improvements over a baseline model using a single-shot prompt on the validation set, while zero-shot prompting offers no measurable gain. We conclude that achieving meaningful improvements through fine-tuning---even with moderate-sized 8B-parameter LLMs and using LoRA---would require computational resources beyond our current budget constraints.

We compare our fine-tuning methods to prompting-based and in-context learning approaches. These include variations on few-shot prompting, self-refinement \cite{madaan2023selfrefine}, self-scoring \cite{bai2022training}, and post-processing.

\subsection{Claimifying Social Media Posts with Self-Refinement \ld{(Nikhil - B4)}}
\label{Claimifying social media posts with self-refinement-Nikhil}

This method uses a combination of prompting strategies and an iterative Self-Refinement stage. In this stage, the same LLM that generates the initial claim provides feedback based on specific criteria that evaluates the check-worthiness of the claim against the input text. This feedback, along with the input and initial claim, is fed back into the same LLM, which generates a refined version of the claim based on the feedback.

We tested both zero-shot and few-shot prompts, with and without a Chain-of-Thought (CoT) trigger phrase. The CoT phrase acts as a cue for the model to reason step by step before producing a final claim. We evaluated these configurations with and without one or more iterations of the self-refinement stage.

For the Task 2 submission, we used the step-by-step "Claimify" process \cite{metropolitansky2025effectiveextractionevaluationfactual}. After extracting the initial claim, we applied one iteration of SELF-REFINE \cite{madaan2023selfrefineiterativerefinementselffeedback}.

We also evaluated a variant using a few-shot prompt followed by the CoT trigger phrase \textprompt{Let's think step by step} \cite{kojima2023largelanguagemodelszeroshot}.

\textbf{Models explored:} GPT-4.1-nano \cite{gpt-4.1-nano-2025-04-14}, Gemini-2.0-Flash \cite{gemini-2.0-flash}, LLaMA-3.3-70B \cite{meta2024llama3.3}, Grok3 \cite{xai2025grok3beta}

\textbf{Highest Avg METEOR score:} 0.332 (Grok3 + Few-shot-CoT)

\textbf{Prompt variations tested:}

\begin{enumerate}
\item Zero-shot: \begin{blockprompt}Identify the decontextualized, stand-alone, and verifiable central claim in the given post:\
\$\{post\}\end{blockprompt}

\item Zero-shot-CoT: Zero-shot + \textprompt{Let's think step by step.}

\item Few-shot: Four examples from the training set followed by the Zero-shot instruction

\item Few-shot-CoT: Few-shot + \textprompt{Let's think step by step.}
\end{enumerate}

\subsection{\ld{Palash - B3} Keyword Few-Shot (KBFP) and Self-Refine}
\label{palash}

This method explores a smart selection of few-shot examples using keyword matching, combined with (or without) a Self-Refine step. All implementations used the LLaMA 3.3 70B model \cite{meta2024llama3.3}.

\paragraph{Keyword Few-Shot.}

The \texttt{Keyword Few-Shot} method selects relevant examples from the training set by matching keywords found in the target social media post. These examples are then used to construct a few-shot prompt \cite{liu2022makes}.

For the example post: \begin{blockexample} The salary of a U.S. Senator is \$174,000 per year. This is Joe Biden's house... seems legit :) \end{blockexample}

The method extracts a claim such as: \begin{blockexample} The main claim is that Joe Biden's house appears to be too expensive for him to afford on a U.S. Senator's salary of \$174,000 per year, implying that there may be some other, potentially questionable, source of income. \end{blockexample}

\paragraph{Self-Refine.}

As an additional step, we apply one iteration of the Self-Refine procedure \cite{madaan2023selfrefineiterativerefinementselffeedback}, using the following prompt: \begin{blockprompt} Refine the following claim to make it more precise.

Here is the text: \$\{the claim\}

Output only the refined claim and nothing else. \end{blockprompt}

While the base Keyword Few-Shot method yields a higher METEOR score on average, we observe that the addition of Self-Refine often produces more concise and less redundant claims.

The refined version of the above claim is: 

\begin{blockexample} Joe Biden's house appears to be too expensive to be affordable solely based on his U.S. Senator's salary of \$174,000 per year, suggesting that there may be an additional, unreported, or unexplained source of wealth that contributed to its purchase or maintenance. \end{blockexample}

We find that repeated applications of Self-Refine do not improve the results. On the contrary, multiple iterations tend to introduce verbosity and hallucinated facts not grounded in the original post.

\paragraph{Issues with the Gold Claim.}

It’s worth noting that the gold-standard claims themselves can have shortcomings. For instance, the gold claim for the example post is: \begin{blockexample} Joe Biden lives in a large estate bought on a senator's salary. \end{blockexample}

This omits key details like the senator's actual salary (\$174,000), which may be necessary for verification and doesn’t reflect the original post’s implication that the estate seems unaffordable based on that salary alone.

\subsection{\ld{(Mitchell - B1)} Subclaim Extraction and Filtering with Refinement}
\label{subclaim-extraction-and-filtering-with-refinement-mitchell}

We explore a multi-stage approach that begins by extracting several potential claims, so-called ``sub claims'', from each social media post. These are then scored \cite{bai2022training} and filtered before a final synthesis step generates the main predicted claim.

In the first stage, LLaMA 3.3 70B model \cite{meta2024llama3.3} is prompted to extsract multiple sub claims from the post. These sub claims represent possible interpretations or factual assertions implied by the content.

Next, we introduce a filtering stage. Rather than passing all sub claims to the claim synthesis step, we rank them using a self-assigned importance score (1 to 10) and retain only those scoring 7 or higher. This limits noise and reduces the cognitive load on the synthesis LLM.

Despite these refinements, the METEOR score did not improve significantly. To address this, we added a third step: post-synthesis revision. A final LLM call revisits the synthesized claim, comparing it with the original post. It performs a "quality check" focused on factual accuracy, emphasis, and eliminating redundancy or verbosity. The prompt in this stage explicitly instructs the model to consolidate language while preserving core meaning.

This three-stage pipeline---subclaim extraction, importance-based filtering, and post-synthesis refinement---aims to balance comprehensiveness with clarity and precision.

\subsection{Max Multi-Prompt \ld{(Benji - B2)}}
\label{benji}

We observed that many social media posts are comments on images found online, while the claims in our dataset often describe those images. If we use a generic prompt, the extracted content won't align well with the dataset's gold claims. To address this, we designed a prompt that instructs the model to imagine searching for the referenced image online and then describe its likely content. A similar idea has also been reported by \citet{perez2021truefewshot}.

We also noted that many posts rely on metaphor or sarcasm, often targeting the government. For instance, when a user writes, ``Biden's annual salary is only \$170K,'' the implication---delivered with irony---is that Biden must be supplementing his income through questionable means to afford a luxury home. Similarly, posts about epidemics often question vaccine policies sarcastically, implicitly accusing public health authorities of negligence or malice.

To account for these nuances, we created targeted prompts tailored to each type of rhetorical device. Empirical results were obtained using the LLaMA 3.3 70B model \cite{meta2024llama3.3}.

This approach demonstrates the potential benefits of intelligently triaging between multiple prompt templates. To simulate an upper bound on this strategy, we applied several different prompts to the same social media post and selected the resulting claim with the highest METEOR score.


\section{Evaluation Approach: Baseline}

\subsection{Regurgitation Baseline \ld{(Aiden - C)}}\label{naive-baseline-aiden}

To evaluate the significance of our METEOR scores, we designed a ``frustratingly  easy'' \cite{daume2007frustratingly} baseline that simply reuses the original social media post or a truncated version as a stand-in for actual claim extraction. Surprisingly, this sets a strong reference point for METEOR performance.

We explored the following variations:

\begin{itemize}
\item Full social media post
\item Truncating after the first 100 characters, omitting partial words at the end
\item Using only the nouns and verbs from the post
\end{itemize}

On the validation set, using the full post yields a METEOR score of 0.19. Truncating the post leads to improved results, with the best configuration achieving 0.24. This baseline scored 0.23 on the test set.

By contrast, using only the nouns and verbs significantly reduces METEOR performance.

\FloatBarrier
\section{Experimental Evaluation}

\subsection{Setup}
\label{setup}

We used only the datasets provided by the CheckThat! Lab Task 2 organizers, focusing exclusively on the English language subset.

Several methods relied on LLMs without additional training, using in-context learning only. In cases where few-shot examples were used, they were selected from the training set. Table \ref{tab:llms} lists which LLM was used in each method that contributed to our empirical results.

For methods involving fine-tuning, all available training data was used.

All approaches were evaluated on the validation subset using the METEOR metric, as implemented in the NLTK toolkit. While we submitted methods with the highest METEOR score, we were interested in exploring methods with the following criteria.

\begin{itemize}
\item Validation METEOR score
\item Novelty and insightfulness of the approach
\item Our subjective preference for the style of the extracted claims
\end{itemize}

\begin{table}[tbp]
\caption{LLM models used in the respective approach. LLMs were either used from HuggingFace or via the Together.AI API service.}
\label{tab:llms}
\begin{tabularx}{\textwidth}{@{}lXl@{}}
\toprule
Section & Method Name & Model \\
\midrule
\ref{finetuned-flan-t5-large-joe} & Finetuned FLAN-T5-Large & FLAN-T5-Large \\
\ref{mohammed} & LoRA fine-tuning of Flan-T5 Base & Flan-T5 Base \\
\ref{adam} & Fine-tuned DeepSeek-R1-Distill-Llama-8b & DeepSeek-R1-Distill-Llama-8b \\
\midrule
\ref{Claimifying social media posts with self-refinement-Nikhil} & Claimifying social media posts with self-refinement & Grok3 \\
\ref{palash} & Self-Refine with KBFP & Llama 3.3 70B \\
\ref{subclaim-extraction-and-filtering-with-refinement-mitchell} & Subclaim extraction and filtering with refinement & Llama 3.3 70B \\
\ref{benji} & Max Multi-Prompt / Single prompt score & Llama 3.3 70B \\
\bottomrule
\end{tabularx}
\end{table}

\begin{table}[tbp]
\caption{Empirical evaluation of our different methods. We focus on METEOR on the validation set and report standard error bars where available. We submitted the three best-performing systems and the baseline to the official leaderboard. Corresponding test set results are shown where available.}
\label{tab:validation}
\begin{tabularx}{\linewidth}{lXccc}
\toprule
Section & Method Name & \multicolumn{2}{c}{METEOR (Validation)} & METEOR (Test) \\
        &             & Score & $\pm$ Error & \\
\midrule
\ref{finetuned-flan-t5-large-joe} 
  & Finetuned FLAN-T5-Large 
  & \textbf{0.5569} & 0.02 & \textbf{0.37} \\

\ref{t5--base-fine-tuned-using-lora-mohammed} 
  & LoRA fine-tuning of FLAN-T5 Base 
  & \multicolumn{2}{l}{0.3054} & 0.28 \\

\ref{adam} 
  & Fine-tuned DeepSeek-R1-Distill-Llama-8b (One-shot) 
  & \multicolumn{2}{l}{0.2541} & \\

\midrule

\ref{Claimifying social media posts with self-refinement-Nikhil}
  & Claimifying social media posts with self-refinement 
  & 0.3310 & 0.007 & 0.33 \\

\ref{palash} 
  & Keyword Based Few-shot Prompt (KBFP) 
  & \multicolumn{2}{l}{0.2943} & \\

\ref{palash} 
  & Self-Refine with KBFP 
  & \multicolumn{2}{l}{0.2392} & \\

\ref{subclaim-extraction-and-filtering-with-refinement-mitchell} 
  & Subclaim extraction and filtering with refinement 
  & 0.2290 & 0.005 & \\

\ref{benji} 
  & Max multi-prompt (simulated) 
  & \multicolumn{2}{l}{0.3277} & \\

\ref{benji} 
  & Single prompt score 
  & \multicolumn{2}{l}{0.2944} & \\

\midrule

\ref{naive-baseline-aiden} 
  & Regurgitation Baseline (full post length) 
  & 0.1944 & 0.008 & \\

\ref{naive-baseline-aiden} 
  & Regurgitation Baseline cut to 100 
  & 0.2429 & 0.008 & 0.23 \\

\bottomrule
\end{tabularx}
\end{table}

\section{Evaluation Results on Validation and Test Set}\label{evaluation-results-on-validation-set}

Table \ref{tab:validation} presents METEOR evaluation results were obtained by our methods on the validation set provided for Task 2. We submitted the best methods and baseline to the challenge, for which test set results are reported as well.

Since Max multi-prompt is a simulated method to explore potential gains, we did not submit it to the challenge.

\paragraph{Discussion of Results.}

The best-performing method was fine-tuning Flan-T5-Large (Section \ref{finetuned-flan-t5-large-joe}), which achieved a METEOR score of 0.5569. This result highlights that larger models do not always guarantee better outcomes---Flan-T5-Large struck the best balance in our experiments.  On the test set, this is still our best approach despite the performance drop.

The runner-up is a prompting-based method based on Claimify with Self-Refine, described in Section \ref{Claimifying social media posts with self-refinement-Nikhil}. It achieved a METEOR score of 0.331 on the validation set, which manifested in a 0.33 on the official test set.

The third-best was T5-Base fine-tuned using LoRA. Although it achieved a lower score of 0.3054, its advantage lies in being a smaller and more efficient model. On the test set, it still obtains a reasonable 0.28 in METEOR.

The fourth method was Keyword-Based Few-Shot Prompting (KBFP), which used few-shot examples with LLaMA 3.3 70B. We did not submit this method to the challenge.

\bigskip

Despite Fine-tuning methods obtaining the highest METEOR scores, we believe that somewhat better claims can be extracted with other approaches.

Subjectively, as we discuss below in Section \ref{examples-of-claims-extracted-on-validation-set}, the most useful claims were generated by one iteration of Self-Refine combined with KBFP or Claimify, particularly for the Joe Biden house example. This approach correctly highlighted the assertion that the house seemed too expensive for his reported salary. In contrast, other methods either focused only on stating the salary amount (\$174,000) or made vague claims about the size of the house. Several outputs hallucinated or speculated about visual content in the image, which was not part of the dataset.

We also observed that the gold claim for this example was not ideal: it omitted both the assertion about affordability and the actual salary figure---both of which are critical for verifying the claim.

\subsection{Overall Leaderboard}

Our best method, described in Section \ref{finetuned-flan-t5-large-joe}, placed us 9th on the leaderboard (Table \ref{tab:leaderboard}).
Notably, rank 12 was occupied by a test submission of the method from Section \ref{Claimifying social media posts with self-refinement-Nikhil}.
Our naive baseline outperformed the final two teams in the rankings.


\begin{table}[tbp]
\caption{Our team ranked 9th on the overall leaderboard.}
\label{tab:leaderboard}
\begin{tabularx}{\textwidth}{@{}rXl@{}}
\toprule
Rank & User & Test Set Results \\
\midrule
1  & tatiana.anikina       & 0.4569 (1) \\
2  & DSGT-CheckThat        & 0.4521 (2) \\
3  & Bharatdeep\_Hazarika  & 0.4114 (3) \\
4  & AKCIT-FN              & 0.4058 (4) \\
\midrule
5  & pratuat.amatya        & 0.4049 (5) \\
6  & rohan\_shankar        & 0.3920 (6) \\
7  & manan-tifin           & 0.3881 (7) \\
8  & MazenYasser74         & 0.3841 (8) \\
\textbf{9}  & \textbf{UNH (Our Team)} & \textbf{0.3737 (9)} \\
\midrule
10 & Ather-Hashmi          & 0.3565 (10) \\
11 & teamopenfact          & 0.3370 (11) \\
12 & Nikhil\_Kadapala      & 0.3321 (12) \\
13 & aryasuneesh           & 0.3153 (13) \\
14 & Soumodeepsahaa        & 0.3098 (14) \\
15 & uhh\_dem4ai           & 0.2612 (15) \\
16 & tomasbernal01         & 0.1660 (16) \\
17 & VSE                   & 0.0070 (17) \\
\bottomrule
\end{tabularx}
\end{table}

\subsection{Examples of Claims Extracted on Validation Set}\label{examples-of-claims-extracted-on-validation-set}

We manually reviewed extracted claims for the first few validation instances. We found considerable variation in claim styles: some were more actionable for fact-checking, while others focused more on the rhetorical tone or the poster's motivation.

Although these differences significantly impact utility for human fact-checkers, they are not well captured by the METEOR metric, which is limited to overlap with gold-standard claims.




\paragraph{ Finetuned Flan-T5-Large.} METEOR score: 0.5569, described in Section \ref{finetuned-flan-t5-large-joe}.

\begin{itemize}
\item
  Extracted Claim: \begin{blockexample}
  Joe Biden owns the house in this photo.
  \end{blockexample}
\item
  Extracted Claim: \begin{blockexample}
  A Holocaust story told in the New York Times
  \end{blockexample}
\item
  Extracted Claim: \begin{blockexample}
      This leopard cub\textquotesingle s mother was killed by a trophy
  hunter
  \end{blockexample}
\item
  Extracted Claim: \begin{blockexample}
  Video shows a crocodile spotted in a residential area in Hyderabad
  during the ongoing heavy rains.
  \end{blockexample}
\end{itemize}

This method was able to closely mimic the patterns of the training dataset in some samples but still failed in other instances. It often does not produce proper claims, as it focuses solely on optimizing the METEOR score.

\paragraph{T5-base (LoRA, fine-tuned)} METEOR Score: 0.3054, described in Section \ref{t5--base-fine-tuned-using-lora-mohammed}.

\begin{itemize}
\item Extracted Claim: \textexample{Joe Biden's house is a fake}\\
\item Extracted Claim: \textexample{Jewish boy adopted by US Jewish family}\\
\item Extracted Claim: \textexample{Video of a crocodile in Hyderabad}\\
\end{itemize}

This model extracts short, concise claims that are easy to verify. However, the claims are often overly literal and fail to capture the deeper meaning or intent behind the social media post.

\paragraph{Fine-tuned DeepSeek-R1-Distill-Lama-8b} METEOR score: 0.2541 , described in Section \ref{adam}.

\begin{itemize}
\item
  Extracted Claim: \begin{blockexample}
  The essential primary claim is that a U.S. Senator earns \$174,000 per
  year, and it is Joe Biden\textquotesingle s house.
  \end{blockexample}
\item
  Extracted Claim: \begin{blockexample}
  The Karnofsky family adopted a 7-year-old boy into their home,
  providing him with food through his homework until he was 12, when
  they helped him buy his first instrument, showcasing their support for
  his musical talent.
  \end{blockexample}
\item
  Extracted Claim: \begin{blockexample}
  The essential primary claim is that trophy hunting is horrific.
  \end{blockexample}
\item
  Extracted Claim: \begin{blockexample}
  The essential primary claim is that none of the listed items
  (Magarmacch, Heavy Rain, Hyderabad, Crocodile, Alert) are the main
  focus.
  \end{blockexample}
\item
  Extracted Claim: \begin{blockexample}
  The administration is now blaming the victims of
  today's deadly attacks in Kabul for not leaving
  earlier.
  \end{blockexample}
\end{itemize}
The extracted claims are generally suitable for fact-checking and, in several cases, elaborate on the underlying message of the post.

We observed that the trained model often prefixes responses with phrases like ``the essential primary claim is...'' This model tends to extract short, concise claims that are easy to verify. However, these claims are frequently overly literal and fail to capture the deeper meaning or intent behind the social media post.

\paragraph{Claimifying social media posts with self-refinement} METEOR score: 0.332, Section \ref{Claimifying social media posts with self-refinement-Nikhil}.

\begin{itemize}
\item 
    Extracted Claim: \textexample{US Senator’s annual salary is \$174,000}
\item 
    Extracted Claim: \begin{blockexample}A Lithuanian Jewish family employed a 7-year-old boy until he was 12 and gave him money to buy his first instrument. \end{blockexample}
\item 
    Extracted Claim: \textexample{Rescued animal's mother was killed by a hunter.}
\item 
    Extracted Claim: \textexample{Crocodile sighted in Hyderabad during heavy rain.}
\end{itemize}

The generated claims are semantically closer to the gold-standard claims. In a separate analysis, we find that an average BERTScore (F1 mean) of 0.82 is achieved. Compared to the original input text, this indicates that the generated claims remain true to both the content and context of the topic, despite achieving a lower METEOR score.

\paragraph{Max multi-prompt.} METEOR score: (up to) 0.3277, described in Section \ref{benji}.
\begin{itemize}
\item
  Extracted Claim: \textexample{The annual salary of a U.S. Senator is \$174,000.}
  
\item
  Extracted Claim: \begin{blockexample}Owning such a home on that salary doesn't add up. This is a tongue-in-cheek critique of perceived wealth versus official pay. \end{blockexample}

\end{itemize}

While the first prompt yields a claim that is objectively fact-checkable, we believe the second prompt better captures the motivation behind the social media post.

\section{Overall Conclusions and Main Findings}

We found that most LLMs produced claims related to the content of the social media posts. However, these often diverged from the gold standard claims, which negatively impacted their METEOR scores. When averaged across 1,170 validation examples, most of our methods converged on a METEOR score of approximately 0.27. The differences between methods only became apparent through manual inspection and checking for coverage of all key claim elements.

We also observed that many gold claims failed to capture all the critical assertions. In longer social media posts, multiple check-worthy claims were often present, making it difficult, even for human judges, to determine the primary claim without additional context about the user's information needs.

Several gold claims referenced images linked in the social media posts. Since these images were not included in the dataset, any claims based on them were speculative.

With sufficient fine-tuning, the smaller FLAN T5 model was able to approximate the method used to extract gold standard claims.

We found that multiple iterations of self-refinement often led LLMs to hallucinate or produce overly verbose responses. This pattern was consistent across a wide range of models, including LLaMA, GPT 4.1 nano, Gemini 2.0 Flash, and Grok3. In some cases, these models returned the same claim without any improvements after a few iterations, which can be attributed to the rigid criterion imposed by the prompt used for both extraction and feedback.

In many cases, the baseline outputs without self-refinement achieved higher METEOR scores and produced claims that more accurately reflected the content of the original post.

We also noted that the instruction following was inconsistent. For example, directives such as omitting phrases like "The main claim is..." were often ignored, particularly by LLaMA models. Using structured outputs, such as JSON or Pydantic formats, improved adherence but frequently resulted in truncated outputs that were no longer valid JSON.

\section*{Acknowledgment and Declaration on Generative AI}

This work was conducted in part by participants of the CS881 graduate course ``Data Science for Knowledge Graph and Text'' at the University of New Hampshire, as well as by students in the Computing Research Association’s UR2PhD program. We gratefully acknowledge their contributions and enthusiasm throughout the research process.

This material is based upon work supported by the National Science Foundation under Grant No. 1846017. Any opinions, findings, conclusions, or recommendations expressed in this material are those of the authors and do not necessarily reflect the views of the National Science Foundation.

During the preparation of this work, the author(s) used ChatGPT, Grammarly in order to: Grammar and spelling check, Paraphrase and reword. After using this tool/service, the author(s) reviewed and edited the content as needed and take(s) full responsibility for the publication’s content.



\bibliography{bib,clef25-checkthat}
\begin{appendices}
\section{Appendix}
\subsection{Prompts used in Claimifying social media posts with Self-refinement}

\subsubsection*{Claim Extraction System Prompt}

\paragraph{Identity / System Message}

\begin{blockprompt}
You are a helpful AI assistant and an expert in claim detection, extraction, 
and normalization.  
\end{blockprompt}

\paragraph{Instructions}

\bigskip

\begin{blockprompt}
\begin{lstlisting}[numbers=none]
 You are given a noisy social media post that contains only text but it might 
  have been posted alongside a photo or video on the platform it was extracted 
  from.
      
* Your task is to detect, extract, and respond with a normalized claim.

* A claim is a statement or assertion that can be objectively verified as true or 
  false based on empirical evidence or reality.
    
* Follow the below steps to analyze the input text and arrive at the final 
  response.
    
* Step 1: Sentence Splitting and Context Creation

    * Start by splitting the post into individual sentences.
    * Now create or retrieve a context for each of those sentences by looking at 
      the two preceding and two following sentences.

* Step 2: Selection

    * Now determine if each sentence contains any verifiable information based 
      on the context created for the sentence in the previous step.
    * For each sentence, do the following:
        * If the sentence does not contain any verifiable information, discard 
          that sentence.
        * If the sentence contains both verifiable and unverifiable information, 
          rewrite the sentence retaining only verifiable information.
        * If the sentence doe not contain any unverifiable information, return 
          the original sentence.
    
* Step 3: Disambiguation

    * The task here is to identify two types of ambiguity. 
    * The first is referential ambiguity, which occurs when it is unclear what a 
      word or phrase refers to. For example, in the sentence "They will update 
      the policy next year", the terms "They," "the policy," and "next year" are 
      ambiguous. 
    * The second is structural ambiguity, which occurs when grammatical structure 
      allows for multiple interpretations. For instance, the sentence "AI has 
      advanced renewable energy and sustainable agriculture at Company A and 
      Company B" can be interpreted as: (1) AI has advanced renewable energy and 
      sustainable agriculture at both Company A and Company B, or (2) AI 
      has advanced renewable energy at Company A, and it has advanced 
      sustainable agriculture at Company B.
    * Now determine whether each instance of ambiguity can be resolved using the 
      question and the context. 
    * The standard for resolution is whether a group of readers would likely 
      agree on the correct interpretation. For example, recall the sentence "AI 
      has advanced renewable energy and sustainable agriculture at Company A and 
      Company B." If the context specified that Company A builds solar panels and 
      Company B reduces farms' water usage, readers would likely conclude that AI 
      has advanced renewable energy at Company A and sustainable agriculture at 
      Company B. Conversely, if the context only described both companies as 
      "environmental pioneers," readers would have insufficient information to 
      determine the correct interpretation.
    * If any ambiguity is unresolvable, discard the sentence even if it has 
      unambiguous, verifiable components.
    * If all ambiguity is resolved, return a clarified version of the sentence. 
    * If there is no ambiguity, retain the original sentence for the next step.
            
* Step 4: Decomposition

    * Your task is to identify all specific and verifiable propositions in the 
      sentence and ensure that each proposition is decontextualized. 
      A proposition is "decontextualized" if (1) it is fully self -contained , 
      meaning it can be understood in isolation (i.e., without the question , the 
      context , and the other propositions), AND (2) its meaning in isolation 
      matches its meaning when interpreted alongside the question , the context , 
      and the other propositions. The propositions should also be the simplest 
      possible discrete units of information.
    
* If no verifiable claims are found, return an extractive summary of the central 
  idea of the post in a single sentence.
    
* Use only the words found in the original input text when generating a response. 
    
* The claim must be strictly extracted from the input without adding any inferred 
  or assumed context.
    
* The claim should be a concise single sentence (up to a maximum of 25 words) 
  that captures the main point of the post without any additional context or 
  details. Prioritize the main claim if multiple claims are present.
    
* The claim should be a self-contained factual statement that can be verified. It 
  should not contain any subjective opinions, speculations, or interpretations.
    
* Pay attention to negative sentiment, named entities, names of people, and 
  linguistic features like assertions, hedges, implications, etc. If any one 
  of these features are present in the post, it should be reflected in the claim.
    
* Do not include any additional information or explanations in your response.

* Minor clarifications (e.g., implied agent) are allowed if they are logically 
  unavoidable and directly inferable from the input.
    
* If the input text contains any Named Entities, they must be included in your 
  responses.
    
* Return your response in the style of a short caption or headline of a news 
  bulletin.
    
* If the given input text is mostly likely to be referencing or directly talking 
  about or posted alongside a photo or video, return the response that starts 
  with either (1) Photo shows <your_response> or (2) Video shows <your_response>.
      
* Always return your response in English even if the original input is in a 
  different language.
\end{lstlisting}
\end{blockprompt}

\textbf{Claim Extraction User Prompt}
\begin{blockprompt}
    
\begin{lstlisting}[numbers=none]
  # Here are some examples of how to identify a decontextualized, stand-alone, 
    and verifiable central claim in a post.
  <user_query id="example-1">
  **Identify the decontextualized, stand-alone, and verifiable central claim in 
  the given post:** Lieutenant Retired General Asif Mumtaz appointed as 
  Chairman Pakistan Medical Commission PMC Lieutenant Retired General Asif 
  Mumtaz appointed as Chairman Pakistan Medical Commission PMC Lieutenant 
  Retired General Asif Mumtaz appointed as Chairman Pakistan Medical Commission 
  PMC None.
  Let's think step by step.
  </user_query>

  <assistant_response id="example-1">
  **Normalized claim:**  Pakistani government appoints former army general to 
  head medical regulatory body.
  </assistant_response>
  ---------------------------------------------

  <user_query id="example-2">
  **Identify the decontextualized, stand-alone, and verifiable central claim in 
  the given post:** A priceless clip of 1970 of Bruce Lee playing Table Tennis 
  with his Nan-chak !! His focus on speed A priceless clip of 1970 of Bruce Lee 
  playing Table Tennis with his Nan-chak !! His focus on speed A priceless clip 
  of 1970 of Bruce Lee playing Table Tennis with his Nan-chak !! His focus on 
  speed None
  Let's think step by step.
  </user_query>

  <assistant_response id="example-2">
  **Normalized claim:**  Late actor and martial artist Bruce Lee playing table 
  tennis with a set of nunchucks.
  </assistant_response>
  ---------------------------------------------

  <user_query id="example-3">
  **Identify the decontextualized, stand-alone, and verifiable central claim in 
  the given post:** Hydrate YOURSELF W After Waking Up Water 30 min Before a Meal 
  DRINK Before Taking a Shower 2192 2192 Before Going to Bed at the correct time 
  T A YE Helps activate internal organs Helps digestion Helps lower blood 
  pressure. Helps to avoid heart attack Health+ by Punjab Kesari.
  Let's think step by step.
  </user_query>

  <assistant_response id="example-3">
  **Normalized claim:** Drinking water at specific times can have different 
  health benefits
  </assistant_response>
  ---------------------------------------------

  <user_query id="example-4">
  **Identify the decontextualized, stand-alone, and verifiable central claim in 
  the given post:** Eating vaginal fluids makes you immune to cancer, and other 
  diseases. Do it for health. Scientists at St. Austin University in North 
  Carolina, they investigated the benefits of vaginal or cervical mucus 
  consumption and the results were amazing. These fluids contain high levels of 
  active proteins up to 10 minutes after leaving the female body. The vaginal 
  fluid is rich in protein, sodium, vitamins like C1, C4, C4, vc and others.
  This study confirms what was exposed by Dr. John d. Moore in his 2009 study of 
  the "equivalent exchange" theory, which indicates that women and men benefit in 
  the same way. The benefits of "swallowing" vaginal fluids are: 
  1. **Eliminates buttons and buttons**  
  2. **Stimulates the electrical charges of the cells**
  3. **Prevents prostate cancer.**
  4. **Improved digestion.**
  5. **Very effective against constipation.**
  6. **It makes teeth and bones stronger.**
  7. **Helps the functioning of the kidneys Share men! Everything is for your health! Share it on all social networks.**
  Let's think step by step.
  </user_query>

  <assistant_response id="example-4">
  **Normalized claim:** St. Austin University North Carolina says eating vaginal 
  fluid makes you immune to cancer
  </assistant_response>
  ---------------------------------------------

  <user_query id="example-5">
  **Identify the decontextualized, stand-alone, and verifiable central claim in 
  the given post:** Corona virus before it reaches the lungs it remains in the 
  throat for four days drinking water a lot and gargling with warm water & 
  salt or vinegar eliminates the virus $\ldots$
  Let's think step by step.
  </user_query>

  <assistant_response id="example-5">
  **Normalized claim:** Gargling water can protect against coronavirus
  </assistant_response>
  ---------------------------------------------

  <|User|>: Identify the decontextualized, stand-alone, and verifiable central 
  claim in the given post: {input} <|End_user|>
  Let's think step by step.
\end{lstlisting}
\end{blockprompt}

\subsubsection*{Feedback Generation Prompt}

\paragraph{Identify / System Message}

\begin{blockprompt}
  You are a professional fact-checker and an expert in claim normalization. 
\end{blockprompt}

\paragraph{Instructions}

\begin{blockprompt}
\begin{lstlisting}[numbers=none]
  Your task is to provide detailed, constructive feedback on the generated
  response based on the criteria provided to ensure that the normalized claims
  are not only consistent with the original post, but are also self-contained and
  verifiable.
  We want to iteratively improve the above generated response. To help with this,
  please score the response on the following criteria using a 0-10 scale, and
  provide a brief justification for each score:

  1. **Verifiability:** To what extent does the response contain claims that can 
     be independently verified using reliable sources? (0 = not verifiable, 
     10 = fully verifiable)
  2. **Likelihood of Being False:** How likely is it that the response contains
     false or misleading information? (0 = very unlikely, 10 = very likely)
  3. **Public Interest:** How likely is the response to be of general public
     interest or relevance? (0 = not interesting, 10 = highly interesting)
  4. **Potential Harm:** How likely is the response to be harmful, offensive, or
     cause negative consequences? (0 = not harmful, 10 = extremely harmful)
  5. **Check-Worthiness:** How important is it to fact-check this response? 
     (0 = not worth fact-checking, 10 = highly worth fact-checking)

  For each criterion, provide:
  - A score (0-10)
  - Provide a short, precise justification in 1 sentence.

  Optionally, suggest specific improvements to the response based on your
  evaluation.
  
  Response/Normalized Claim: ${Extracted Claim}
\end{lstlisting}
\end{blockprompt}

\subsubsection*{Refined Claim Generation Prompt}

\bigskip

\paragraph{Identity / System Message}

\begin{blockprompt}
  You are a professional fact-checker and expert in claim normalization. 
\end{blockprompt}

\paragraph{Instructions}

\bigskip

\begin{blockprompt}
\begin{lstlisting}
  * Your task is to refine the genrated response in light of the feedback
    provided.
  * Using the feedback provided, return a refined version of the generated
    response, ensuring that the normalized claim is consistent with the original
    post, self-contained, and verifiable.
  * Your response must only be based on the feedback provided.
  * Do not speculate, provide subjective opinions, or add any additional
    information or explanations. 
  * Only include the refined, normalized claim in your response. 
  * If no meaningful refinement is necessary, re-output the original normalized
    claim as-is.
  * If the response is not not decontextualized, stand-alone, and verifiable,
    improve the response by adding more context from the original post if needed.
    
  <|user_query|>{original user query}<|end_of_user_query|>
  <|assistant_response|>{Initial Claim}<|end_of_assistant_response|>
  <|feedback|>{feedback}<|end_of_feedback|>
  <|instruction|>Based on the feedback provided, please refine the above
    generated response/normalized claim.<|end_of_instruction|>
  
\end{lstlisting}
\end{blockprompt}
\end{appendices}

\end{document}